\documentclass[runningheads]{llncs}
\usepackage{graphicx}
\usepackage{amsmath}
\usepackage{booktabs}

\usepackage{multirow}
\usepackage{makecell}
\usepackage{array}
\newcolumntype{P}[1]{>{\centering\arraybackslash}p{#1}}
\newcommand\mr[2]{\multicolumn{1}{c}{\multirow{#1}{*}{\makecell{#2}}}}

\usepackage[inline]{enumitem}

\usepackage{xspace}

\DeclareMathAlphabet{\mathsl}{OT1}{ptm}{m}{sl}

\newcommand{\myagent}{MyAgent\xspace}
\newcommand{\opagent}{OpAgent\xspace}

\usepackage{xcolor}

\begin{document}
\title{A Data-Driven Method for Recognizing Automated Negotiation Strategies}
%
%
\author{Ming Li\inst{1} \and
Pradeep K.Murukannaiah\inst{2} \and
Catholijn M.Jonker\inst{2}}
\authorrunning{Ming et al.}
%
\institute{University of Amsterdam 
\email{uestcliming@gmail.com}\\
\and
Delft University of Technology\\
\email{\{P.K.Murukannaiah,C.M.Jonker\}@delft.nl}}
\maketitle              
\begin{abstract}
Understanding an opponent agent helps in negotiating with it. Existing works on understanding opponents focus on preference modeling (or estimating the opponent's utility function). An important but largely unexplored direction is recognizing an opponent's negotiation \emph{strategy}, which captures the opponent's tactics, e.g., to be tough at the beginning but to concede toward the deadline. Recognizing complex, state-of-the-art, negotiation strategies is extremely challenging and simple heuristics may not be adequate for this purpose.

We propose a novel data-driven approach for recognizing an opponent's negotiation strategy. Our approach includes
\begin{enumerate*}[label=(\arabic*)]
    \item a data generation method for an agent to generate domain-independent sequences by negotiating with a variety of opponents across domains,
    \item a feature engineering method for representing negotiation data as time series with time-step features and overall features, and
    \item a hybrid (recurrent neural network based) deep learning method for recognizing an opponent's strategy from the time series of bids.
\end{enumerate*}
We perform extensive experiments, spanning four problem scenarios, to demonstrate the effectiveness of our approach.

\end{abstract}
\section{Introduction}
Negotiation is a joint decision making process, wherein participants seek to reach a mutually beneficial agreement. It is a core activity in human society and widely exists in social and organizational settings. Automated negotiation \cite{fatima2014principles} involves intelligent agents negotiating on behalf of humans, aiming to not only save time and effort for humans but also yield better outcomes than human negotiators \cite{bosse2005human}. Automated negotiation can play an important role in application domains, including supply chain, smart grid, digital markets, and autonomous driving.

In a negotiation, the better you understanding the opponent's negotiation strategy the easier it is to reach win-win outcomes, reduce negotiation costs, and avoid exploitation by the opponent \cite{baarslag2016learning}. Accordingly, there has been an emphasis on \emph{opponent modeling}. Two key aspects of an opponent model are the opponent's
\begin{enumerate*}[label=(\arabic*)]
    \item \emph{preference profile}, capturing \emph{what} the opponent wants, and 
    \item \emph{negotiation strategy}, capturing \emph{how} the opponent negotiates to achieve a preferred outcome.
\end{enumerate*}
Preference modeling has received considerable attention in the literature. For example, an effective and simple strategy for preference modeling while negotiating is Smith's frequency model \cite{van2012agent}, evaluated in \cite{hendrikx2012evaluating}.
Recognizing an opponent's negotiation strategy is still an unsolved problem (as discussed further in Section~\ref{sec:related-works}).

In recent years, a number of complex, well-performing, negotiation strategies have been developed. For example, several well-performing strategies feature in the annual Automated Negotiation Agents Competition (ANAC) \cite{baarslag2012first}. Further, the GENIUS \cite{hindriks2009genius} repository hosts several state-of-the-art strategies.

\begin{figure}[!htb]
\centering
\includegraphics[width=0.5\columnwidth]{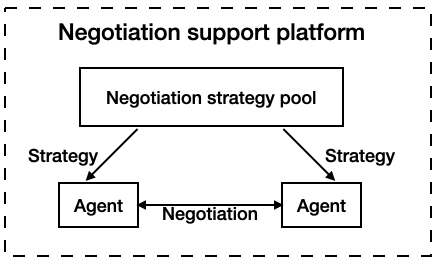}
\caption{The architecture of a simple negotiation support platform}
\label{fig:negotiation-platform}
\end{figure}

Designing a well-performing negotiation strategy is nontrivial. Most end users of automated negotiation, e.g., manufacturers, retailers, and customers on an ecommerce platforms such as Alibaba and Ebay, may not be able to develop their own negotiation strategy but can benefit from strategies designed by professionals. To better serve such users and increase the adoption of automated negotiation, a negotiation support platform, which provides a \emph{strategy pool} for users to choose a strategy from, could be designed (Figure~\ref{fig:negotiation-platform}). However, the performance of a strategy is highly dependent on the opponents and the negotiation scenarios; no single strategy is always ideal. Thus, different people would choose different strategies based on their personal experience.


We make the first attempt at recognizing complex, well-performing, (including state-of-the-art), negotiation strategies via a data-driven method. We explore this challenging problem in a simplified but important setting, where agents select their strategies from a strategy pool. That is, the objective of our agent (detector) is to recognize which strategy the opponent is adopting from the pool. If our approach yields promising results, we open a novel research avenue on modeling opponent strategies, including opportunities to relax simplifying assumptions (as discussed further in Section~\ref{sec:future}).


We propose a data-driven approach for opponent strategy recognition. In our approach, the agent learns to recognize strategies from a dataset of \emph{negotiation traces}. A negotiation trace is the sequence of bids the negotiators make in a negotiation \cite{hindriks2011let}. Based on a history of past negotiations, the agent can learn a classification model of negotiation strategies. Using that model, and based on the negotiation trace up to the point of the latest bid, the agent can classify the opponent's strategy, and adapt accordingly.

We perform extensive experiments to evaluate the effectiveness of our approach. We select eight state-of-the-art (ANAC finalists) and two basic negotiation strategies in our experimental strategy pool. We include four domains of different sizes, and opponents of different preference profiles, resulting in different opposition (competitiveness) during negotiation. We organize our experiments along four problem scenarios, varying the amount of information an agent may have about the domain and the opponent (answer to each can be none). We measure the accuracy of opponent strategy recognition at different time points in a negotiation.

\subsubsection{Contributions}
\begin{enumerate}
\item We propose a data generation and feature engineering method for curating a rich, domain-independent, dataset of negotiation time series, and a hybrid neural network model for opponent strategy recognition. To the best of our knowledge, ours is the first work on recognizing an opponent's complex negotiation strategy. 

\item We provide empirical evidence that strategy recognition is feasible,  in our simplified case. This opens up a research line in strategy recognition, for more complex situations with a bigger pool of known and unknown opponents. 

\end{enumerate}

\section{Related Works}
\label{sec:related-works}

Automated negotiation dates back to the 1980's when ecommerce took flight, e.g., \cite{sycara1988resolving}. The field was formalized in the 1990's (e.g., \cite{rosenschein1994rules,sandholm1999automated}). The need for rigorous means for evaluation of the quality of negotiating agents led to metrics \cite{lomuscio2003classification}, and later to the opensource negotiation platform GENIUS to enable benchmarking \cite{hindriks2009genius}, and the annual ANAC (Automated Negotiation Agents Competition) in 2010 \cite{baarslag2012first}. By now, GENIUS holds a host of agents (including the best from ANAC), negotiation domains and preference profiles. 

\subsection{Opponent Strategy Recognition}

Besides understanding the outcome preferences of an opponent, predicting the opponent's negotiation behaviour would benefit negotiators. 
There are several works on forecasting an opponent's future bids, e.g., \cite{brzostowski2006adaptive,carbonneau2011pairwise}.

\cite{matos1998determining} employ an evolutionary method to determine successful bidding strategies, where genes represent the parameters of the bidding strategy. This method can be used to optimize different classes of negotiation strategies, but does not help to determine which class is best for a given negotiation. \cite{fatima2001optimal} mathematically optimize negotiation strategies for different circumstances, concluding that further optimization is possible, once we have information on the negotiation strategy of the opponent. Thus, recognizing the opponent strategy matters, subscribed by Harverd Business School, e.g., \cite{barron2007negotiation}.

\cite{lai2010learning} learn behavioral regularities of the opponent using fuzzy constraints and based on these regularities learn the opponent's concession rate. Later \cite{chen2013conditional} did the same using Boltzmann machines. 
For single issue negotiation in bilateral negotiation \cite{papaioannou2011multi} compare the performance of multiple estimators to predict the opponent's bidding strategy. Finally, \cite{koeman2019recognising} focus on basic strategies and a setting in which the negotiators only have 10 rounds to negotiate. They show that it is possible to recognize the opponent's basic strategy by logical analysis of the opponent's behavior. In short, earlier attempts focus on abstract aspects of bidding strategies and not on recognizing specific (complex) strategies.

\subsection{Strategy Recognition in Other Fields}
In the broader field of agent technology, case-based plan recognition (CBPR) comes close to the strategy recognition challenge. In CBPR, each encountered sequence of actions is assigned a support count, which is used to identify common strategies in a given game, e.g., \cite{fagan2003case}. However, as \cite{weber2009data} remark, CBPR does not scale well to real-time problems with an increase in the number and complexity of possible action sequences. Note that for a general negotiation system strategy recognition is even more difficult as it  has to deal with the fact that the domain of negotiation can change per negotiation. This is also why the ideas of \cite{weber2009data} are not applicable. Once a number of strategy recognizing systems have been created, the approach of \cite{alserhani2010mars}, from cybersecurity research, to recognize attack strategies might become applicable.

\section{Approach}
We describe our negotiation setting, the strategy recognition problem, and the three steps of our approach. 

\subsection{Preliminaries}
\subsubsection{Setting.}
Suppose that an agent, \myagent{} ($M$) and an opponent agent, \opagent ($O$), engage in a bilateral negotiation. The agents follow the well-known alternating offers protocol \cite{rubinstein1982perfect}, where an agent starts with a offer (or bid, $b$); then on, each agent can accept or reject the other agent's offer (ending the negotiation), or make a counter offer (continuing the negotiation). For simplicity (without loss of generality), we assume that \myagent{} starts the negotiation. A negotiation can last several rounds; each round $i$ (except the last) consists of two offers $\langle b_M^i, b_O^i\rangle$ (the last round will have only one offer if \myagent{} ends negotiation). An example negotiation trace, for a negotiation that lasts 10 rounds which \opagent{} ends, is $\{\langle b_M^1, b_O^1\rangle, \langle b_M^2, b_O^2\rangle, \ldots, \langle b_M^{10}, b_O^{10}\rangle\}$. 

Further, suppose that there is a pool of well existing negotiation strategies, $S = \{S_1, S_2, \ldots, S_n\}$. The \opagent{} employs a strategy in $S$ throughout a negotiation. We do not require \myagent{}'s negotiation strategy to be in $S$, since we regard \myagent{} as a detector agent whose characteristics will be discussed in the following section.

\subsubsection{Dataset.}
Suppose that \myagent{} maintains a dataset of negotiation traces, where each trace has a label from $S$, indicating the opponent's strategy corresponding to that trace. The dataset includes multiple traces for each opponent strategy in $S$. We assume that \myagent{}'s strategy is the same across all these traces. If \myagent{} employs different strategies, it needs to maintain a dataset for each strategy. \myagent{} can maintain such datasets by simulating negotiations and by including traces from real negotiations it participates in.

\subsubsection{Problem.}
Suppose that \myagent{} and \opagent{} are in a negotiation, where \opagent{} employs a negotiation strategy from $S$, and \myagent{} maintains a dataset of negotiation traces including strategies from $S$. Let it be the beginning of round $i$, when it is \myagent{}'s turn to place a bid. The trace for the negotiation so far is $t = \{\langle b_M^1, b_O^1\rangle, \ldots, \langle b_M^{i}, b_O^{i}\rangle\}$. 

Then, our problem is, given the dataset $D$ and trace $t$, how can \myagent{} recognize \opagent{}'s negotiation strategy?
We seek to answer this question in four scenarios shown in Table~\ref{tab:problem-scenarios}, depending on \myagent{}'s knowledge about the domain of negotiation and \opagent{}'s preference profile.

\begin{table}[!htb]
\centering
\caption{Four scenarios for the opponent recognition problem}
\label{tab:problem-scenarios}
\begin{tabular}{@{}P{1cm}P{2.8cm}P{3.6cm}@{}}
\toprule
Problem\newline scenario & \myagent domain \newline experienced? & \myagent's knowledge of\newline \opagent's preferences \\
\midrule
P1 & Yes & Complete \\
P2 & Yes & Partial \\
P3 & Yes & None \\
P4 & No & None \\
\bottomrule
\end{tabular}
\end{table}

We consider \myagent as experienced in a domain if it has negotiated in that domain (for real or in simulations), so that $D$ includes negotiation traces from that domain. We consider \myagent's knowledge of \opagent's preference as
\begin{enumerate*}[label=(\arabic*)]
    \item complete, if \myagent knows the exact preference profile of \opagent (which can be the case in some repeated negotiations);
    \item partial, if \myagent has encountered \opagent{} in the past but the uncertainty of the estimated opponent profile is low or if the domain is partially predictable \cite{aydougan2018machine}; and
    \item none, if \myagent has not negotiated with the opponent before and the domain is not predictable.
\end{enumerate*}

\subsection{Overview of our approach}
\begin{figure}[!htb]
\centering
\includegraphics[width=0.75\columnwidth]{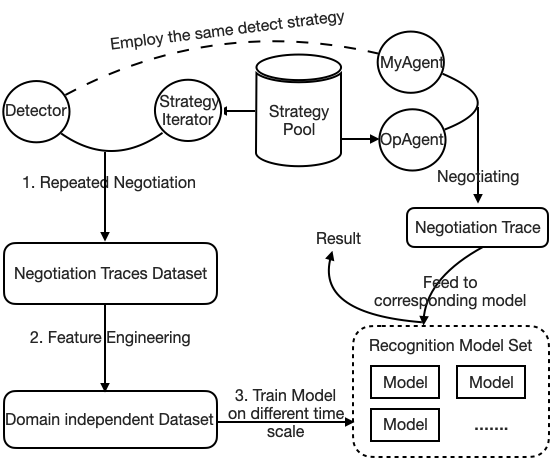}
\caption{The overall approach of strategy recognition.}
\label{fig:overall-approach}
\end{figure}

Figure~\ref{fig:overall-approach} shows our overall approach which consists of three key steps: data generation, feature engineering, and training. 

In the first step, we prepare a negotiation trace to be suitable as a training instance for a learning algorithm. An agent can simulate a dataset of negotiation traces via repeated negotiation with a strategy negotiator of the strategy pool. 

In the second step, we engineer features from the negotiation traces for a time series. We transform the domain-specific bids to domain-independent utilities so that an agent can
\begin{enumerate*}[label=(\arabic*)]
    \item learn from traces across domains, and
    \item predict strategies in a domain the agent was not trained on.
\end{enumerate*}

Finally, in the third step, we train several hybrid deep learning models to recognize negotiation strategies from the time series. The models are trained on different negotiation rounds and together form a recognition model set.

In the recognition phase, the agent can employ the same detection strategy as one of the trained models and generate negotiation traces as the negotiation goes. By feeding the negotiation trace to the corresponding model the agent gets recognition results.

\subsection{Data Generation}
To generate the dataset, we fix \myagent{}'s negotiation strategy. Although \myagent{} can employ any negotiation strategy, we desire the strategy to be
\begin{enumerate*}[label=(\arabic*)]
    \item not easily conceding, so that \myagent{} does not end the negotiation too fast; and
    \item not too tough, so that \opagent{} does not walk away from the negotiation;
\end{enumerate*}
For the experiments we report on in this paper, we let \myagent{} employ the nice tit-for-tat strategy.

Next, we select negotiation domains, and preference profiles for \myagent{} and \opagent{}, depending on the problem scenario. Section~\ref{subsec:experimental-settings} provides concrete examples used in our experiments. Then, we assign a strategy from the strategy pool $S$ to \opagent{}, and simulate several negotiations between \myagent{} and \opagent{}. After each negotiation, we add the negotiation trace and the opponent strategy label to the dataset. A negotiation trace consists of patterns of offers and counteroffers. We represent a negotiation trace as a time series, preserving the sequential information. We repeat this process for each strategy in $S$.

\subsection{Feature Engineering}
In order to learn patterns from sequences, we represent each bid trace as a time series, considering each bidding round as a time step. Then, we engineer domain-independent features from domain-dependent bids in the time series.

First, we compute utilities from bids. Let $U_M$ and $U_O$ be the actual utility functions of \myagent{} and \opagent{}. The agents may not know each other's actual utility functions but can estimate each other's utility functions. Let $\widehat{U}_M$ and $\widehat{U}_O$ be the estimated utility functions of \myagent{} and \opagent{}. For concreteness, we employ the Smith Frequency Model (SFM) \cite{van2012agent}, which employs frequencies of issue values in a negotiation to estimate utilities. Given a bid sequence $\omega$ from an agent, the SFM utility function is:

\begin{align}
\label{eq:SFM}
    \widehat{U}(\omega) &= \left (  {\sum_{i\in I} } \widehat{w}_{i} \times \widehat{e}_{i}(\omega_{i}) \right ) \times {}\frac{1}{{\sum_{i \in I}}{\widehat{w}_{i}}},~~\text{where} \\
    \widehat{w}_{i} &= \frac{C^{+}_{i}}{C^{sum}_{i}},~~\text{and}~~ \widehat{e}_{i}(\omega_{i})=\frac{C_{\omega_{i}}}{C^{+}_{i}},~~\text{where} \nonumber \\
    C^{+}_{i} &= \max_{v\in V_{i}}C_{v}~~\text{and}~~C^{sum}_{i} = \sum_{v \in V_{i}}C_{v},~~\text{where} \nonumber
\end{align}
\noindent $V_{i}$ is the value set of issue $i$; $C_{v_{i}}$ is the count of value $v_{i}$ in the received offers; $C^{+}_{i}$ and $C^{sum}_{i}$ are the maximum and the sum of value counts for values of issue $i$, respectively; $\widehat{w}_{i}$ and $\widehat{e}_{i}$ are the estimated weight and the estimated evaluation function for issue $i$; $\omega_{i}$ is the sub-outcome for issue $i$.

Baarslag et al. \cite{baarslag2013predicting} did an elaborate comparison of the methods to estimate opponent utility and found that \emph{frequency models} and \emph{value models} have good performance. Considering the performance and simplicity, we choose SFM. Further, we focus only on the linear additive profiles. In future work, we can study more complex profiles. 

Given four utility functions and two bids in each time step, we compute eight utility-based features shown in Table~\ref{tab:basic-features}. Note that even if we have my own utility $U_m(b_m)$, we still include the estimated utility $\widehat{U}_m(b_m)$, since the opponent could also have a opponent model, that in turn influences its behavior.

\begin{table}[!htb]
\centering
\caption{The basic features computed for each negotiation round}
\label{tab:basic-features}
\begin{tabular}{c@{\hspace{.4cm}}l@{\hspace{.4cm}}c@{\hspace{.4cm}}c}
\toprule
& & \multicolumn{2}{c}{Bid} \\
\cmidrule(lr){3-4}
& & \myagent & \opagent \\
\midrule
\parbox[t]{6mm}{\multirow{4}{*}{\rotatebox[origin=c]{90}{\shortstack{Utility\\Function}}}}
 & \myagent (Actual) & $U_m(b_m)$ & $U_m(b_o)$ \\ 
 & \opagent (Actual) & $U_o(b_m)$ & $U_o(b_o)$ \\ 
 & \myagent (Estimated) & $\widehat{U}_m(b_m)$ & $\widehat{U}_m(b_o)$ \\ 
 & \opagent (Estimated) & $\widehat{U}_o(b_m)$ & $\widehat{U}_o(b_o)$ \\ 
\bottomrule
\end{tabular}
\end{table}

\myagent{} may not be able to employ all eight features in each problem scenario (Table~\ref{tab:problem-scenarios}). Specifically, in scenarios P2, P3 and P4, \myagent{} does not know \opagent{}'s actual utility function. In those scenarios, we do not employ features depending on \opagent{}'s actual utility function.

We include two additional type of features, considering utility changes from one negotiation round to the next. 

First, we consider the amount of difference for each basic utility function, e.g., for \myagent{}'s actual utility function, the change in utility for round $i$ is: $U_m^i - U_m^{i-1}$. 

Second, we apply DANS analysis \cite{hindriks2011let} to derive analytical information from the utility changes. Let $\Delta_m^i = U_m(b_o^i) - U_m(b_o^{i-1})$, and $\Delta_o^i = U_o(b_o^i) - U_o(b_o^{i-1})$. In computing $\Delta_o^i$, if \myagent{} does not know $U_o$, it can employ $\widehat{U}_o$, instead. Then, via DANS analysis, we categorize \opagent{}'s move ($b_o^{i-1} \rightarrow b_o^{i}$) as one of the following.

\begin{itemize}[noitemsep]
    \item Fortunate: $\Delta_o^i > \gamma$, $\Delta_m^i > \gamma$;
    \item Selfish: $\Delta_o^i > \gamma$, $\Delta_m^i < -\gamma$;
    \item Concession: $\Delta_o^i < -\gamma$, $\Delta_m^i > \gamma$;
    \item Unfortunate: $\Delta_o^i < -\gamma$, $\Delta_m^i < -\gamma$;
    \item Nice:  $\Delta_o^i \in [-\gamma, \gamma]$, $\Delta_m^i > \gamma$;
    \item Silent:  $\Delta_o^i \in [-\gamma, \gamma]$, $\Delta_m^i \in [-\gamma, \gamma]$.
\end{itemize}
In our experiments, we set $\gamma$ to 0.002, which works well for our pool of negotiation strategies and scenarios. Also, we encode DANS categories via one-hot encoding, representing each category as a binary-valued feature. Finally, we compute overall features considering the entire time series. The overall features include the
\begin{enumerate*}[label=(\arabic*)]
    \item basic utilities from the last round;
    \item change in utilities from the first to the last round;
    \item sum of each DANS category across all time steps; and
    \item round at which the negotiation ends.
\end{enumerate*}

Table~\ref{tab:feature-summary} summarizes the features we engineer. Note that scenarios P2, P3, and P4 have fewer features than P1 since we do not compute features based on \opagent{}'s actual utility function in problem scenarios P2, P3, and P4.

\begin{table}[!htb]
\centering
\caption{Summary of the feature we employ}
\label{tab:feature-summary}
\begin{tabular}{@{}lc@{\hspace{.4cm}}c@{\hspace{.4cm}}c@{}}
 \toprule
  \multirow{2}{*}{Feature} & \multirow{2}{*}{Type} & \multicolumn{2}{c}{Count} \\
  \cmidrule{3-4}
  & & P1 & P2, P3, P4 \\
 \midrule
  Utilities & Time-step & 8 & 6 \\
  Change in utilities & Time-step & 8 & 6 \\
  DANS category & Time-step & 6 & 6 \\
  Last round utilities & Overall & 8 & 6 \\
  Change in utilities (first to last) & Overall & 8 & 6 \\
  Sum of DANS categories & Overall & 6 & 6 \\
  Number of negotiation rounds & Overall & 1 & 1 \\
 \bottomrule
\end{tabular}
    \label{tab:my_label}
\end{table}

\subsection{Strategy Recognition Model}

\begin{figure}[!htb]
\centering
\includegraphics[width=0.75\textwidth]{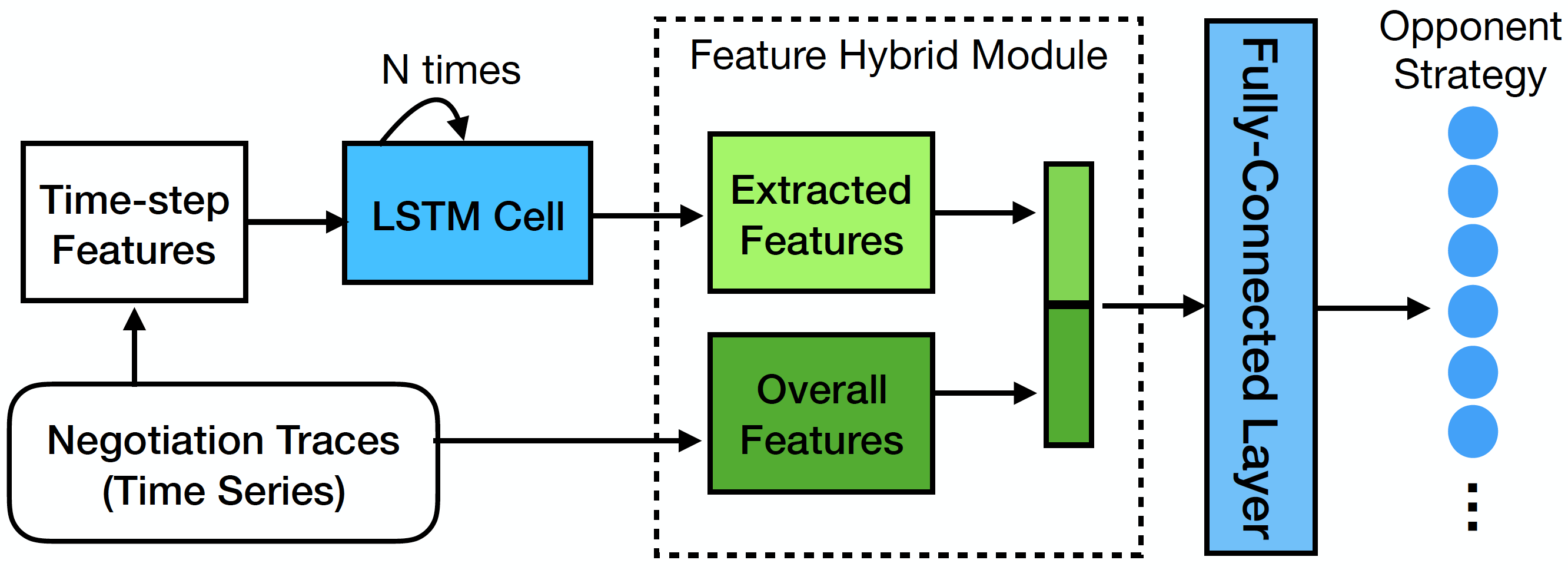}
\caption{The opponent strategy recognition model's architecture}
\label{fig:model-architecture}
\end{figure}

To recognize the opponent strategy from the strategy pool, we propose a novel hybrid deep learning model.  Figure~\ref{fig:model-architecture} depicts the overall architecture of the model, consisting of three main modules: LSTM-based recurrent neural network, hybrid-feature module, and a fully-connected layer. 

Recurrent neural network (RNN) architecture is well-suited for modeling sequential data. Thus, in our recognition model, the LSTM \cite{hochreiter1997long} layer takes negotiation time-series data (with its time-step features) as input and automatically extracts valuable features. We extract 64 features.

The hybrid-feature module combines the features extracted from LSTM layer and overall features to form a hybrid feature representation. Finally, the fully-connected layer maps the hybrid features into opponent strategies. We apply the softmax function to get the probability of the classification results. We use cross-entropy loss function, which is widely used for classification. In the training phase, network parameters are optimized to minimize the loss.

We can train different models to recognize opponent strategies at different negotiation rounds. We choose the number of LSTM cells, accordingly, e.g., to recognize strategy after 20 rounds, we set $N=20$.

\section{Experiments}
\label{subsec:experimental-settings}
We describe, first, the general experimental settings, and then, four experiments, one for each problem scenario.


\subsection{General experimental settings}
\subsubsection{Opponent Strategy Pool.}
We use 10 strategies---two basic strategies: RandomCounteroffer, Boulware; and eight well performed complex startegies from ANAC finalists: DoNA (2014), Atlas3 (2015), Fairy (2015), Caduceus (2016), YXAgent (2016), CaduceusDC16 (2017), GeneKing (2017), Rubick (2017). Each strategy has different characteristics and outperforms the others in several situations. For instance, DoNA is a domain-based negotiator approach using behavioral strategies, which employs a cognitive model to divide the class of all possible domains into different regions based on the analysis regarding the time and concession stance. Caduceus and Caduceus16 use a meta strategy to collect opinions of a set of negotiation expert strategies.
Geneking uses Genetic Algorithm to explore negotiation outcome space.

\subsubsection{Domain and Preference Profiles.}
We select four domains from ANAC 2015 of varying size $\langle$number of issues, size of the outcome space$\rangle$: Bank Robbery (Bank) $\langle$3, 18$\rangle$, Car Profile (Car) $\langle$4, 240$\rangle$, University (Uni) $\langle$5, 11250$\rangle$, and Tram $\langle$7, 972$\rangle$.
In each domain, \myagent{} sticks to a randomly chosen preference profile. We vary \opagent{}'s preferences profile, to include different levels of \emph{opposition} \cite{baarslag2013evaluating} between the agents. Conceptually, opposition indicates the extent to which the profiles are (not) compatible---the higher the opposition, the higher the competitiveness of the negotiation. We do not include combinations of preference profiles that lead to a low competitive negotiations (e.g., when both agents have the same preference). In such cases, the negotiation is likely to end quickly and recognizing opponent strategy may not be necessary.

\subsubsection{Platform.}
In each domain, for each opponent strategy and each combination of preference profiles, we simulate 50 negotiations on GENIUS, with 100 rounds as deadline.

\subsubsection{Model Setting.} We use the same general training setting for each model. We use is the Adam optimizer, with a learning rate of 0.001, $\beta_{1}$ = 0.5, $\beta_{2}$ = 0.999, and the training batch size is set to 64. In all experiments, the recognition model is trained for 80 epochs (unless it converges sooner).

In each experiment, we split the data 80-20 for training and testing, and measure the model accuracy.

\subsection{P1: One Domain and Known Preferences}

\begin{figure}[!b]
\centering
\includegraphics[width=1\textwidth]{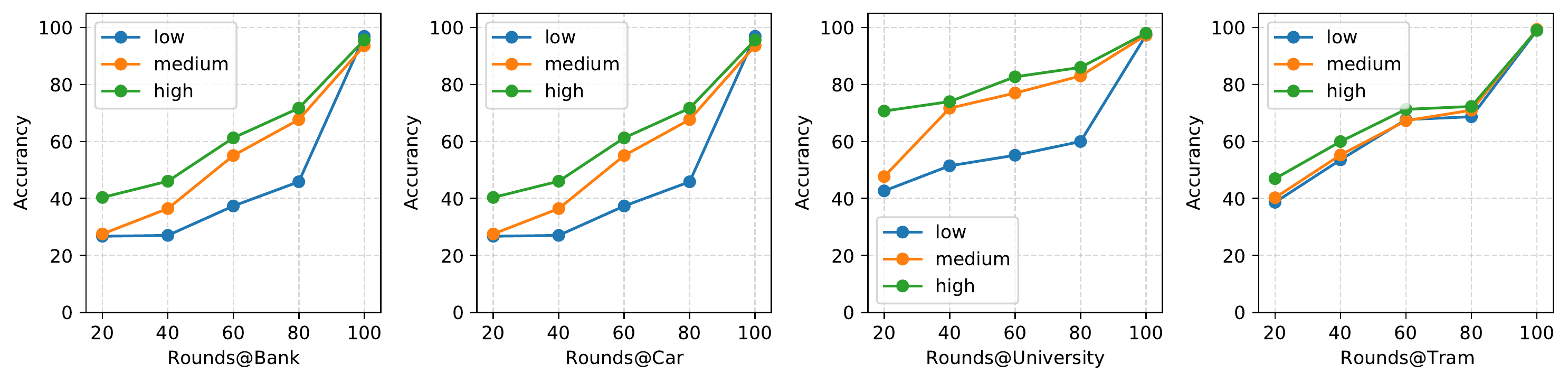}
\caption{The influence of competitiveness (Experiment P1)}
\label{fig:known-preference}
\end{figure}

P1 is the basic case, where \myagent{} knows the domain and \opagent{}'s preference. Both sides put their preferences on the table during the negotiation or the preference of a specific role is common sense or predictable, i.e., buyer and seller negotiation. This setting serves as simplest case and the baseline of the strategy recognition problem. In this case, our model employs all information in the dataset, i.e., 22 features as input to LSTM module and 87 (64 LSTM-extracted plus 23 overall) features as input to the fully-connected layer. Also, this scenario serves the upper bound on the accuracy our model can yield.

In this experiment, we also evaluate the influence of the opposition between agents' profiles on model accuracy. To do so, we select three opponent preference profiles, resulting in opposition values in ranges; 0.1--0.2, 0.2--0.3, $>$ 0.3.

We make three key observations from the P1 results shown in Figure~\ref{fig:known-preference}. First, overall, our model yields better accuracy than random guessing (whose accuracy would be 10\% since we have a 10-class balanced classification problem). Further, the model accuracy increases for longer traces, which is expected since longer sequences are likely to include more information. The high accuracy with complete sequences (100 rounds) suggests that, in a repeated negotiation, \myagent{} can estimate \opagent{}'s strategy in the first negotiation and employ it for adapting strategy in subsequent negotiations.

Second, we observe that there is a huge improvement in recognition accuracy when given the traces of 100 rounds instead of 80 rounds. We conjecture that this is mainly because most finalist negotiation strategies tend to perform in a tough manner \cite{jonker2017automated}, so the last few bids of a negotiation are the most influential ones.

Third, we find an interesting pattern about the influence of opposition: the higher the opposition, the better the model accuracy. We conjecture that an \opagent{} does not ``exert'' as much strategy in a low-competitive negotiation as it does in a high-competitive negotiation. Accordingly, traces from competitive negotiations are more indicative of an \opagent{}'s strategy. In competitive negotiations, the model accuracy is relatively high even in the earlier rounds.

\subsection{P2: One Domain and Known Preference Set}
In P2, we assume that \opagent{}'s preference belongs to a fixed preference set (which \myagent{} could have estimated), but we don't know which one exactly. Thus, we do not include features based on \opagent{}'s utility $U_{O}$. We input 16 features to LSTM and 83 (64 + 19) features to the fully-connected layer (this setting applies to P2, P3, and P4).

In this experiment, we also explore the influence of domain size on accuracy. Since the opposition can influence accuracy (as the previous experiment shows), we control the average opposition for each domain to be 0.2 $\pm$ 0.005.

As Table~\ref{tab:known-preference-sets} shows, P2 shows a similar pattern as P1 with respect to the increasing accuracy over negotiation rounds and the significant improvement from 80 to 100 rounds. However, the accuracy drops from P1 to P2 since we have less information about the opponent in P2 than in P1.

\begin{table}[!htb]
\centering
\caption{The influence of domain size $\langle$number of issues, size of outcome space$\rangle$ on model accuracy for known preference sets (P2)
}
\label{tab.domaininfluence}
\label{tab:known-preference-sets}
\begin{tabular}{@{}l@{\hspace{.3cm}}l@{\hspace{.3cm}}l@{\hspace{.3cm}}l@{\hspace{.3cm}}l@{\hspace{.3cm}}l@{\hspace{.3cm}}l@{}}
\toprule
\multirow{2}{*}{Domain} & \multirow{2}{*}{Size} & \multicolumn{5}{c}{Negotiation Rounds} \\ 
\cmidrule(l){3-7} 
& & 20 & 40 & 60 & 80 & 100\\
\midrule
Bank & $\langle$3, 18$\rangle$ & 26.9 & 32.7 & 42.9 & 50.6 & 81.4\\
Car & $\langle$4, 240$\rangle$ & 35.4 & 42.6 & 61.3 & 69.8 & 93.1\\
Uni & $\langle$5, 11250$\rangle$ & 53.8 & 63 & 71.2 & 72.9 & 94.2\\
Tram & $\langle$7, 972$\rangle$ & 39.7 & 59.6 & 71.3 & 73 & 98.7\\
\midrule
Average & & 39 & 49.5 & 61.7 & 66.6 & 91.9\\
\bottomrule
\end{tabular}
\end{table}


We observe that the domain size influences model accuracy: the bigger the domain the higher the accuracy, in general. We conjecture that an \opagent{} has more room to exercise its strategy in a bigger domain than in a small domain, making it easier to recognize strategies in bigger domains. 
Both domain size variables seem to influence model accuracy. Specifically, the University domain has more issues but smaller outcome space than the Tram domain. The model accuracy for Tram, compared to University, is higher at later rounds (60, 80, 100) but lower at earlier rounds (20, 40).

\subsection{P3: One Domain and Unknown Preferences}

In P3, we assume that \myagent{} does not know \opagent{}'s preference. We train our model using observations from an \opagent{} of one preference profile but test the model on an \opagent{} of a different profile. Further, to understand the influence of opposition, we test a given trained model against multiple \opagent{}s, varying the opposition values.

Table~\ref{tab:unknown-preference} shows the model accuracy at 100 and 60 negotiation rounds. The accuracy drops from P2 to P3, as expected. Importantly, this suggests that using strategy prediction with preference estimation (as in scenario P2) is valuable.

\begin{table}[!htb]
\centering
\caption{Accuracy when \opagent{}'s preference is unknown (P3)}
\label{tab:unknown-preference}
\begin{tabular}{@{}c@{}l@{\hspace{0.5cm}}l@{\hspace{0.5cm}}l@{\hspace{0.5cm}}l@{}} 
\toprule
\mr{2.87}{Domain (Train\\opposition)} & \multicolumn{4}{c}{Accuracy (Test opposition)} \\ 
\cmidrule(lr){2-5}
 & \multicolumn{4}{c}{Negotiation Round $=$ 100} \\
\cmidrule(lr){1-1} \cmidrule(lr){2-5}
Uni (.17) & 35.9 (.07) & 64.8 (.17) & 45.6 (.25) & 57.5 (.32) \\ 
Tram (.18) & 53.7 (.06) & 77.3 (.15) & 88.9 (.19) & 60.1 (.27) \\
Bank (.24) & 41.4 (.04) & 56.1 (.18) & 35.7 (.25) & 34.1 (.36) \\
Car (.28) & 82.5 (.11) & 65.5 (.22) & 86.4 (.29) &  88.1 (.32) \\
\cmidrule(lr){2-5}
& \multicolumn{4}{c}{Negotiation Round $=$ 60} \\
\cmidrule(lr){2-5}
Uni (.17) & 23.3 (.07) & 32.4 (.17) & 36.1 (.25) & 51.2 (.32) \\
Tram (.18) & 45.1 (.06) & 59.8 (.15) & 60.9 (.19) & 51.3 (.27) \\
Bank (.24) & 21 (.04) & 19.3 (.18) & 18.8 (.25) & 19.2 (.36) \\
Car (.28) & 56.5 (.11) & 56.7 (.22) & 50.3 (.29) &  40 (.32) \\
\bottomrule
\end{tabular}
\end{table}

We did not find a clear connection between the opposition values of profiles in the training and test sets, and model accuracy. At 100 rounds, a model trained with profiles of a certain opposition yields highest accuracy for test profiles of a similar opposition (although this is not the case for Bank domain). However, at 60 rounds, the best performing test profiles' opposition value is not necessarily similar to the training profiles' opposition value. We need better mechanisms (than relying solely on the opposition values) for building a dataset for training a model to recognize strategies in scenarios where an \opagent{}'s preference is completely unknown.

\subsection{P4: Cross Domain}
In P4, we train our model on one domain and test it on another. We control the opposition values of the train and test profiles to be similar ($\sim$0.18).

Table~\ref{tab:cross-domain} shows the cross-domain accuracy at 100 and 60 negotiation rounds. P4 is the most challenging scenario for our strategy recognition approach. Yet, we observe that the accuracy of our model is better than random guessing. 

We observe that the accuracy varies significantly for different train-test domain pairs. For example, at 100 rounds, the model trained on the University domain yields an accuracy of 72.1\% when tested on the Tram domain, but the accuracy for Bank (training) and University (test) is only 39.7\%. Similarly, at 60 rounds, Car and Tram work with each other (as train-test pairs) much better than other pairs.
\begin{table}[!htb]
\centering
\caption{Cross-domain accuracy (P4)}
\label{tab:cross-domain}
\begin{tabular}{@{}c@{\hspace{0.5cm}}c@{\hspace{0.5cm}}c@{\hspace{0.5cm}}c@{\hspace{0.5cm}}c@{\hspace{0.5cm}}c@{\hspace{0.5cm}}c@{\hspace{0.5cm}}c@{\hspace{0.5cm}}l@{}} 
\toprule
\mr{3.87}{Test\\ Domain} & \multicolumn{8}{c}{Train Domain} \\
\cmidrule(l){2-9}
 & \multicolumn{4}{c}{Nego. Round $=$ 100} & \multicolumn{4}{c}{Nego. Round $=$ 60} \\ 
\cmidrule(l){2-5} \cmidrule(l){6-9} 
 & Bank & Car & Uni & Tram & Bank & Car & Uni & Tram \\
\cmidrule(lr){1-1} \cmidrule(lr){2-5} \cmidrule(lr){6-9} 
Bank & -- & 48.3 & 44.7 & 42.5 & -- & 29.4 & 22.7 & 29.2 \\ 
Car &  52.3& -- & 55.4 & 46.3 & 29.1 & -- & 21.9 & 52 \\
Uni & 39.7 & 49.4 & -- & 58 & 23.4 & 22.1 & -- & 23.4 \\
Tram & 45.7 & 49.7 & 72.1 & -- & 38.4 & 57.6 & 25 & -- \\ 
\bottomrule 
\end{tabular}
\end{table}

\vspace{-1em}
\subsection{Experiment Discussion}

In general, the tougher the opposition, the higher the number of issues and the bigger the outcome space, the better the accuracy of our model. The only way one can differentiate between negotiation strategies is if these strategies behave differently. In small domains, the negotiation strategies have less option to choose from, e.g., the number of possible bids in the Bank domain is only 18. Now, suppose that these bids have strict preference ordering. When making 100 bids the strategy can only make 17 concessions, which means that all other 83 bids are repetitions of earlier bids. In that light it is easy to see that even a strictly conceding agent is hard to differentiate from a hardheaded strategy.

Our analysis suggests that transferring models across domains is challenging. However, choosing the right combination of train and test domains can yield higher accuracy. A better understanding of the compatibility of domains and its influence on model accuracy would require an analysis with more domains, considering different domain characteristics.

As stated above, we can only differentiate between strategies if they behave differently. The confusion matrices we computed show that indeed it is easier to differentiate between strategies that make fundamentally different choices. In particular, we found that our model can still recognize the RandomStrategy with high confidence, even in simple scenarios (small domains, few of issues, and low opposition of the preferences). The RandomStrategy is rather unique in how it explores the outcome space. 

\section{Research Directions}
\label{sec:future}

Our results show that strategy recognition is feasible, but there is room for improvement (e.g., higher accuracy in earlier rounds), even in the simplified setting. Besides, there are novel directions to study the strategy recognition.

\subsubsection{Generalized Setting}
A more generalized setting for strategy recognition is when an opponent employs a strategy not in the pool. Recognizing strategies that an agent has never met before is extremely challenging. 
A possible direction is to \emph{cluster strategies}, capturing high-level features to get strategy archetypes, to create a pool of strategy archetypes. However, clustering strategies will be a challenging task. For example, current strategies employ many advanced techniques including meta-learning, genetic algorithm, and cognitive models. In addition, some strategies take opinions from other complex strategies, leading to a fuzzy boundary between archetypes. Another possible direction is to recognize specific characteristics of a strategy, e.g., whether a strategy is cooperative or not.

\subsubsection{Repeated Negotiations}
If we recognize an opponent, we can select and adapt our strategy to achieve a better outcome. In repeated negotiations, the agent could learn to recognize the opponent over multiple sessions, then select a suitable strategy from the pool to achieve higher individual utility or social welfare in later  sessions.

\subsubsection{Strategy Concealment}
As strategy recognition approaches mature, it is likely that opponents conceal their strategy, making detection hard. How can an \opagent conceal its strategy, and how can \myagent detect an opponent's strategy when concealing tactics are in place, are both exciting directions for future work.


\section{Conclusion}
To our knowledge, we make the first attempt at solving the strategy recognition problem. Our data-driven approach includes systematic steps to generate data, feature engineering, and training a hybrid RNN based model.
We conjecture that our approach for data generation and feature engineering can be employed to address negotiation-related problems besides strategy recognition.


We evaluate our approach in four settings. In the simplest setting (one domain, known preferences), our approach yields an accuracy of  up to 83\%, at 60 (out of 100) rounds depending on the domain and opposition. In the most complex setting (cross domain), our approach yields an accuracy of up to 58\% at 60 rounds. However, there is a large variance in accuracy; in some cases the accuracy is as low as 20\%. Characterizing the settings under which strategy recognition yields effective results requires a deeper understanding of the domains, and the agents' preferences and strategies. 



We demonstrate that opponent strategy recognition is a promising research line. Our work opens up novel opportunities to, e.g., cluster strategies, employ counter-detection tactics, and provide negotiation support.

\bibliographystyle{splncs04}
\bibliography{ref}





\end{document}